\documentclass[9pt]{article}

\usepackage[final]{neurips_2020}

\usepackage[utf8]{inputenc} 
\usepackage[T1]{fontenc}    
\usepackage{hyperref}       
\usepackage{url}            
\usepackage{booktabs}       
\usepackage{amsfonts}       
\usepackage{amsthm}         
\usepackage{nicefrac}       
\usepackage{microtype}      
\usepackage{caption}
\usepackage{mathtools}

\newtheorem*{definition}{Definition}

\usepackage{bm}
\usepackage{amsmath}
\usepackage{amssymb}
\usepackage{graphicx}

\usepackage{tikz}
\usetikzlibrary{%
  arrows,%
  backgrounds,
  intersections,
  shapes,
  shapes.misc,
  shapes.arrows,%
  chains,%
  matrix,%
  positioning,
  scopes,%
  decorations.pathmorphing,
  shadows,%
  patterns
}

\graphicspath{{./plots/}}
\usepackage{pgfplots}
\pgfplotsset{compat=newest}
\usepgfplotslibrary{groupplots}
\usepgfplotslibrary{polar}
\usepgfplotslibrary{smithchart}
\usepgfplotslibrary{statistics}
\usepgfplotslibrary{dateplot}

\newcommand{\real}{\text{Re}}

\newcommand{\Wre}{\bm W_{r}}
\newcommand{\Wim}{\bm W_{i}}
\newcommand{\npu}{\text{NPU}}
\newcommand{\naivenpu}{\text{NaiveNPU}}
\newcommand{\nau}{\text{NAU}}

\newcommand{\dense}{\text{Dense}}
\newcommand{\chain}{\text{Chain}}
\newcommand{\norm}[1]{\left\lVert #1 \right\rVert}

\usepackage{booktabs}
\renewcommand{\arraystretch}{1.2}

\title{Neural Power Units}
\author{
  Niklas Heim,
  Tom\'a\v s Pevn\'y,
  V\'aclav \v Sm\'idl\\
  Artificial Intelligence Center\\
  Czech Technical University\\
  Prague, CZ 120 00\\
  \texttt{\{niklas.heim, tomas.pevny, vasek.smidl\}@aic.fel.cvut.cz}\\
}

\begin{document}

\maketitle

\begin{abstract}
  Conventional Neural Networks can approximate simple arithmetic operations,
  but fail to generalize beyond the range of numbers that were seen during
  training.  \emph{Neural Arithmetic Units} aim to overcome this difficulty,
  but current arithmetic units are either limited to operate on positive
  numbers or can only represent a subset of arithmetic operations. We introduce
  the \emph{Neural Power Unit} (NPU).\footnote{
    Implementation of Neural Arithmetic Units: \url{github.com/nmheim/NeuralArithmetic.jl}
    The code to reproduce our experiments is available at \url{github.com/nmheim/NeuralPowerUnits}.
  }
  that operates on the \emph{full domain of
  real numbers} $\mathbb{R}$ and is capable of \emph{learning arbitrary power functions} in
  a single layer.  The NPU thus fixes the shortcomings of existing arithmetic
  units and extends their expressivity. We achieve this by using
  complex arithmetic without requiring a conversion of the network to complex
  numbers $\mathbb{C}$.
  A simplification of the unit to the \emph{RealNPU} yields a highly transparent model.
  We show that the NPUs outperform their competitors in terms of
  accuracy and sparsity on artificial arithmetic datasets, and that the RealNPU can discover
  the governing equations of a dynamical system only from data.
\end{abstract}

\section{Introduction}%
\label{sec:introduction}

Numbers and simple algebra are essential not only to human intelligence but
also to the survival of many other
species~\citep{dehaene_number_2011,gallistel_finding_2018}. A successful,
intelligent agent should, therefore, be able to perform simple arithmetic.  State
of the art neural networks are capable of learning arithmetic, but they
fail to extrapolate beyond the ranges seen during
training~\citep{suzgun_evaluating_2018,lake_generalization_2018}.  The
inability to generalize to unseen inputs is a fundamental problem that hints at
a lack of \emph{understanding} of the given task. The model merely memorizes
the seen inputs and fails to abstract the true learning task.  The failure of
numerical extrapolation on simple arithmetic tasks has been shown
by~\cite{trask_neural_2018}, who also introduced a new class of \emph{Neural
Arithmetic Units} with good extrapolation performance on some arithmetic
tasks.

Including Neural Arithmetic Units in standard neural networks promises to
significantly increase their extrapolation capabilities due to their inductive
bias towards numerical computation. This is especially important for tasks in which
the data generating process contains mathematical relationships.  They also
promise to reduce the number of parameters needed for a given
task, which can improve the explainability of the model.
We demonstrate this in a \emph{Neural Ordinary Differential Equation} (NODE,
\cite{chen_neural_2019}), where a handful of neural arithmetic units can
outperform a much bigger network built from dense layers (Sec.~\ref{sub:fractional_sir_model}).
Moreover, our new unit can be used to directly read out the correct generating ODE
from the fitted model.  This is in line with recent efforts to build
\emph{transparent} models instead of attempting to explain black-box models
\citep{rudin_stop_2019}, like conventional neural networks.
We refer to the terminology by \cite{lipton_mythos_2017} which defines the
potential of understanding the parameters of a given model as \emph{transparency by decomposability}.

The currently available arithmetic units all have different strengths and
weaknesses, but none of them solve simple arithmetic completely.
The \emph{Neural Arithmetic Logic Unit} (NALU) by \cite{trask_neural_2018},
chronologically, was the first arithmetic unit. It can solve
addition ($+$, including subtraction), multiplication ($\times$), and division
($\div$), but is limited to positive inputs. The convergence of the
NALU is quite fragile due to an internal gating mechanism between addition and
multiplication paths as well as the use of a logarithm which is problematic for
small inputs.  Recently, \cite{schlor_inalu_2020} introduced the \emph{improved NALU} (iNALU,
to fix the NALU's shortcomings. It significantly increases its complexity, and we
observe only a slight improvement in performance.
\cite{madsen_neural_2020} solve ($+,\times$) with two new units: the
\emph{Neural Addition Unit} (NAU), and the \emph{Neural Multiplication Unit}
(NMU). Instead of gating between addition and multiplication paths, they are
separate units that can be stacked. They can work with the full range of real
numbers, converge much more reliably, but cannot represent
division.

\subsection*{Our Contributions}%
\label{sub:our_contribution}

\textbf{Neural Power Unit.}
We introduce a new arithmetic layer (NPU, Sec.~\ref{sec:neural_power_unit})
which is capable of learning products of power functions $(\prod x_{i}^{w_i})$ of
arbitrary real inputs $x_i$ and power $w_i$, thus including multiplication
($x_1\times x_2 = x_1^1 x_2^1$) as well as division $(x_1\div
x_2=x_1^{1}x_2^{-1})$. This is achieved by using formulas from complex
arithmetic (Sec.~\ref{sub:naive_npu}).  Stacks of NAUs and NPUs can thus learn
the full spectrum of simple arithmetic operations.

\textbf{Convergence improvement.}
We address the known convergence issues of neural arithmetic units by
introducing a \emph{relevance gate} that smooths out the loss surface of the
NPU (Sec.~\ref{sub:npu}). With the relevance gate, which helps to learn to
ignore variables, the NPU reaches extrapolation errors and sparsities that are on
par with the NMU on ($\times$) and outperforms NALU on ($\div,\sqrt{\cdot}$).

\textbf{Transparency.}
We show how a power unit can be used as a highly transparent\footnote{as defined by
\cite{lipton_mythos_2017}} model for equation discovery
of dynamical systems. Specifically, we demonstrate its ability to identify
a model that can be interpreted as a SIR model with fractional powers
(Sec.~\ref{sub:fractional_sir_model}) that was used to fit the COVID-19
outbreak in various countries \citep{taghvaei_fractional_2020}.

\section{Related Work}%
\label{sec:related_work}

Several different approaches to automatically solve arithmetic tasks have
been studied in recent years. Approaches include Neural GPUs
\citep{kaiser_neural_2016}, Grid LSTMs \citep{kalchbrenner_grid_2016}, Neural
Turing Machines \citep{graves_neural_2014}, and Neural Random Access Machines
\citep{kurach_neural_2016}. They solve tasks like binary addition and
multiplication, or single-digit arithmetic. The Neural Status Register
\citep{faber_neural_2020} focuses on control flow.  The Neural Arithmetic
Expression Calculator \citep{chen_neural_2018}, a hierarchical reinforcement
learner, is the only method that solves the division problem, but it operates on
character sequences of arithmetic expressions. Related is symbolic integration
with transformers \citep{lample_deep_2019}. Unfortunately, most of the named models
have severe problems with extrapolation \citep{madsen_measuring_2019,saxton_analysing_2019}.
A solution to the extrapolation problem could be Neural Arithmetic Units. They are
designed with an inductive bias towards systematic, arithmetic computation.
However, currently, they are limited in their capability of expressing the full
range of simple arithmetic operations ($+,\times,\div$).
In the following two sections, we briefly describe the currently available
arithmetic layers, including their advantages and drawbacks.

\subsection{Neural Arithmetic Logic Units}%
\label{sub:neural_arithmetic_logic_unit}

\citet{trask_neural_2018} have demonstrated the severity of the extrapolation
problem of dense networks for even the simplest arithmetic operations, such as
summing or multiplying two numbers.  To increase the power of
abstraction for arithmetic tasks, they propose the \emph{Neural Arithmetic
Logic Unit} (NALU), which is capable of learning ($+,\times,\div$).  However,
the NALU cannot handle negative inputs correctly due to the logarithm in
Eq.~\ref{eq:nalu_mult}:

\begin{definition}[{\bf NALU}]
  The NALU consists of a ($+$) and a ($\times$) path with shared weights $\bm W$ and $\bm M$.
  \begin{align}
    \label{eq:nalu_add}
    \text{Addition: }       & \bm a = \bm{\hat W} \bm x
                            & \bm{\hat W}& = \tanh(\bm{W}) \odot \sigma(\bm{M}) \\
    \label{eq:nalu_mult}
    \text{Multiplication: } & \bm m = \exp (\bm{\hat W}\log(|\bm x|+\epsilon)) & &\\
    \text{Output: }         & \bm y = \bm a \odot \bm g + \bm m \odot (1-\bm g) 
                            & \bm g& = \sigma(\bm G\bm x)
  \end{align}
  with inputs $\bm x$ and learnt parameters $\bm W$, $\bm M$, and $\bm G$.
\end{definition}

Additionally, the logarithm destabilizes training to the extent that the chance of success
can drop below 20\% for ($+,\times$), it becomes practically impossible to
learn ($\div$) and difficult to learn from small inputs in general
\citep{madsen_measuring_2019}.
\cite{schlor_inalu_2020} provide a detailed description of the shortcomings of
the NALU, and they suggest an \emph{improved NALU} (iNALU).  The iNALU addresses
the NALU's problems through several mechanisms.  It has independent
addition and multiplication weights for Eq.~\ref{eq:nalu_add} and
Eq.~\ref{eq:nalu_mult}, clips weights and gradients to improve training
stability, regularizes the weights to push them away from zero, and, most
importantly, introduces a mechanism to recover the sign that is lost due to the
absolute value in the logarithm. Additionally, the authors propose to
reinitialize the network if its loss is not improving during training.  We
include the iNALU in one of our experiments and find that it only slightly
improves the NALU's performance (Sec.~\ref{sub:simple_arithmetic_task}) at the
cost of a significantly more complicated unit. Our NPU avoids all these mechanisms
by internally using complex arithmetic.

\subsection{Neural Multiplication Unit \& Neural Addition Unit}%
\label{sub:neural_multiplication_unit}

Instead of trying to fix the NALU's convergence issues,
\cite{madsen_neural_2020} propose a new unit for ($\times$) only.
The \emph{Neural Multiplication Unit} (NMU) uses explicit multiplications and learns
to gate between identity and ($\times$) of inputs. The NMU is
defined by Eq.~\ref{eq:nmu} and is typically used in conjunction with the
so-called \emph{Neural Addition Unit} (NAU) in Eq.~\ref{eq:nau}.

\begin{definition}[{\bf NMU \& NAU}]
  NMU and NAU are two units that can be stacked to model ($+,\times$).
  \begin{align}
    \label{eq:nmu}
    \text{NMU: } &y_j = \prod_i \hat M_{ij} x_{i} + 1 - \hat M_{ij}  & \hat M_{ij}=\min(\max(M_{ij}, 0), 1)\\
    \label{eq:nau}
    \text{NAU: } &\bm y = \bm{\hat{A}} \bm x & \hat A_{ij}=\min(\max(A_{ij}, -1), 1)
  \end{align}
  with inputs $\bm x$, and learnt parameters $\bm M$ and $\bm A$.
\end{definition}

Both NMU and NAU are regularized with $\mathcal{R} = \sum_{ij} \min(|W_{ij}|,
|1-W_{ij}|)$, and their weights are clipped, which biases them towards learning
an operation or pruning it completely.  The combination of NAU and NMU can thus
learn $(+,\times)$ for both positive and negative inputs. Training NAU and NMU
is stable and succeeds much more frequently than with the NALU, but they cannot
represent ($\div$), which we address with our NPU.

\section{Neural Power Units}%
\label{sec:neural_power_unit}

To fix the deficiencies of current arithmetic units, we propose a new
arithmetic unit (inspired by NALU) that can learn arbitrary products of power functions $(\prod x_{i}^{w_i})$ (including
$\times,\div$) for positive and negative numbers, and still train well.
Combined with the NAU, we solve the full range of arithmetic operations.
This is possible through a simple modification of the ($\times$)-path of the
NALU (Eq.~\ref{eq:nalu_mult_2}). 
We suggest to replace the logarithm of the absolute value by the complex logarithm and
to allow $\bm W$ to be complex as well. Since the complex logarithm is defined for
negative inputs, the NPU does not have a problem with negative numbers. A complex $\bm W$ 
improves convergence at the expense of transparency (see
Sec.~\ref{sub:fractional_sir_model}). The improvement during training might be
explained by the additional imaginary parameters that make it possible to avoid
regions with an uninformative gradient signal.

\subsection{Naive Neural Power Unit -- NaiveNPU}%
\label{sub:naive_npu}

With the modifications introduced above we can extend the multiplication path
of the NALU from
\begin{align}
  \label{eq:nalu_mult_2}
  \bm m = \exp (\bm W\log_{\text{real}}(|\bm x|+\epsilon))
\end{align}
to use the complex logarithm ($\log\coloneqq\log_{\text{complex}}$) and a
complex weight $\bm W$ to
\begin{align}
  \label{eq:1}
  \bm y = \exp(\bm W\log \bm x) = \exp\left((\Wre + i\Wim) \log\bm x\right),
\end{align}
where the input $\bm x$ is still a vector of real numbers.
With the polar form for a complex number $z=re^{i\theta}$ the complex log applied
to a real number $x=re^{ik\pi}$ is
\begin{align}
  \log x = \log r + ik\pi,
\end{align}
where $k=0$ if $x\geq0$ and $k=1$ if $x<0$.
Using the complex log in Eq.~\ref{eq:1} lifts the positivity constraint on $\bm x$,
resulting in a layer that can process both positive and negative numbers correctly.
A complex weight matrix $\bm W$ somewhere in a larger network would result in complex
gradients in other layers.  This would effectively result in doubling the
number of parameters of the whole network. As we are only interested in real
outputs, we can avoid this doubling by considering only the real part
of the output $\bm y$:
\begin{align}
  \real(\bm y) &= \real(\exp((\Wre + i\Wim)(\log\bm r + i\pi\bm k))) \\
    \label{eq:npumult}
    &= \exp(\Wre\log\bm r - \pi\Wim\bm k) \odot \cos(\Wim\log\bm r + \pi\Wre\bm k).
\end{align}
Above we have used Euler's formula $e^{ix} = \cos x + i\sin x$.
A diagram of the NaiveNPU is shown in Fig.~\ref{fig:npu_diagram}.

\begin{definition}[{\bf NaiveNPU}]
  The Naive Neural Power Unit, with matrices $\Wre$ and
  $\Wim$ representing real and imaginary part of the complex numbers, is defined as
  \begin{gather}
    \label{eq:naivenpu_def}
    \bm y = \exp(\Wre \log\bm r - \pi\Wim\bm k) \odot \cos(\Wim\log \bm r + \pi\Wre\bm k), \text{ where }\\
    \nonumber
    \bm r = |\bm x| + \epsilon,
    \quad
    k_i = \begin{cases}
       0 & x_i \geq 0 \\
       1 & x_i < 0
    \end{cases},
  \end{gather}
  with inputs $\bm x$, machine epsilon $\epsilon$, and learnt parameters $\Wre$ and $\Wim$.
\end{definition}

\begin{figure}
  \begin{minipage}{.45\textwidth}
      \resizebox{0.9\textwidth}{!}{
%
%

\definecolor{c1}{HTML}{8097e9}
\definecolor{c2}{HTML}{c6dea2}
\definecolor{c3}{HTML}{ffb200}
\definecolor{c4}{HTML}{b6b6b6}

\tikzset{
  function/.style={
    rectangle,
    rounded corners=4,
    minimum height=6mm,
    thick,
    draw=black,         
    top color=c1,              
    bottom color=c1, 
    font=\tt
  },
  vector/.style={
    rounded rectangle,
    minimum size=6mm,
    thick,draw=black,
    top color=c2!80,
    bottom color=c2!80,
    font=\ttfamily},
  weight/.style={
    rounded rectangle,
    minimum size=6mm,
    thick,draw=black,
    top color=c3!80,
    bottom color=c3!80,
    font=\ttfamily},
  skip loop/.style={to path={-- ++(0,#1) -| (\tikztotarget)}}
}

{
  \tikzset{vector/.append style={text height=1.5ex,text depth=.25ex}}
  \tikzset{function/.append style={text height=1.5ex,text depth=.25ex}}
}

\begin{tikzpicture}[
        point/.style={coordinate},>=stealth',thick,draw=black!90,
        tip/.style={->,shorten >=0.007pt},every join/.style={rounded corners},
        fat/.style={-, ultra thick, opacity=0.5},every join/.style={rounded corners},
        hv path/.style={to path={-| (\tikztotarget)}},
        vh path/.style={to path={|- (\tikztotarget)}},
        text height=1.5ex,text depth=.25ex 
    ]
    \fill [c4, rounded corners=20, draw, opacity=0.5] (-4.7,-2.0) rectangle (4.5,2.0);
    \matrix[column sep=2mm, row sep=1mm] {
      & & \node (r) [vector] {$\bm r$};&
        \node (log) [function] {log}; & &
        \node (wr1) [function] {matmul}; & & \\

      & \node (abs) [function] {abs}; & & & & &
        \node (wi1) [function] {matmul};& & \node (plus) [function] {$\bm -$}; & &
        \node (exp) [function] {exp};\\

      \node (in) [vector] {$\bm x$}; & \node (p2) [point] {};&
      \node (clip) [point] {}; & & &
      \node (wr) [weight] {$\bm W_r$}; & \node (wi) [weight] {$\bm W_i$}; & & & & &
      \node (mul) [function] {$\bm\odot$}; & &
      \node (out) [vector] {$\bm y$};\\

      & \node (0pi) [function] {0:$\pi$}; & & & & & 
        \node (wi2) [function] {matmul};& & \node (minus) [function] {$\bm +$};& &
        \node (cos) [function] {cos};\\

      & & \node (k) [vector] {$\bm k$}; & &
          \node (beforewr2) [point] {}; & 
        \node (wr2) [function] {matmul}; & & & \\
    };

    { [start chain]
      \chainin (in);
      \chainin (p2) [join];
      { [start branch=rbr]
        \chainin (abs) [join];
        \chainin (r) [join=by {vh path}];
        \chainin (log) [join=by tip];
        { [start branch=rbrup]
          \chainin (wr1) [join=by tip];
          \chainin (plus) [join=by {tip, hv path}];
        }
        { [start branch=rbrdown]
          \chainin (wi2) [join=by {vh path, tip}];
          \chainin (minus) [join=by tip];
        }
        \chainin (plus);
        \chainin (exp) [join=by tip];
        \chainin (mul) [join=by {hv path,tip}];
      }
      { [start branch=kbr]
        \chainin (0pi) [join];
        \chainin (k) [join=by {vh path}];
        \chainin (beforewr2) [join];
        { [start branch=kbrdown]
          \chainin (wr2) [join=by tip];
          \chainin (minus) [join=by {hv path,tip}];
        }
        { [start branch=kbrup]
          \chainin (wi1) [join=by {vh path,tip}];
          \chainin (plus) [join=by tip];
        }
        \chainin (minus);
        \chainin (cos) [join=by tip];
        \chainin (mul) [join=by {hv path,tip}];
        \chainin (out) [join=by tip];
      }
    }

    { [start chain]
      \chainin (wr2);
      \chainin (wr) [join=by fat];
      \chainin (wr1) [join=by fat];
    }

    { [start chain]
      \chainin (wi2);
      \chainin (wi) [join=by fat];
      \chainin (wi1) [join=by fat];
    }

\end{tikzpicture}

}
      \captionof{figure}{NaiveNPU diagram, with input $\bm x$ and output $\bm y$.
      Vectors in green, trainables in orange, functions in blue.}
      \label{fig:npu_diagram}
  \end{minipage}
  \hspace{0.01\textwidth}
  \begin{minipage}{0.54\textwidth}
    \resizebox{\textwidth}{!}{
%
%

\definecolor{c1}{HTML}{8097e9}
\definecolor{c2}{HTML}{c6dea2}
\definecolor{c3}{HTML}{ffb200}
\definecolor{c4}{HTML}{b6b6b6}

\tikzset{
  function/.style={
    rectangle,
    rounded corners=4,
    minimum height=6mm,
    thick,
    draw=black,         
    top color=c1,              
    bottom color=c1, 
    font=\tt
  },
  vector/.style={
    rounded rectangle,
    minimum size=6mm,
    thick,draw=black,
    top color=c2!80,
    bottom color=c2!80,
    font=\ttfamily},
  weight/.style={
    rounded rectangle,
    minimum size=6mm,
    thick,draw=black,
    top color=c3!80,
    bottom color=c3!80,
    font=\ttfamily},
  skip loop/.style={to path={-- ++(0,#1) -| (\tikztotarget)}}
}

{
  \tikzset{vector/.append style={text height=1.5ex,text depth=.25ex}}
  \tikzset{function/.append style={text height=1.5ex,text depth=.25ex}}
}

\begin{tikzpicture}[
        point/.style={coordinate},>=stealth',thick,draw=black!90,
        tip/.style={->,shorten >=0.007pt},every join/.style={rounded corners},
        fat/.style={-, ultra thick, opacity=0.5},every join/.style={rounded corners},
        hv path/.style={to path={-| (\tikztotarget)}},
        vh path/.style={to path={|- (\tikztotarget)}},
        text height=1.5ex,text depth=.25ex 
    ]
    \fill [c4, rounded corners=20, draw, opacity=0.5] (-6.8,-2.0) rectangle (6.6,2.0);
    \fill [c4, draw, pattern=north west lines, pattern color=c4] (-4.9,-2.0) rectangle (-0.9,2.0);
    \matrix[column sep=2mm, row sep=1mm] {
      & & \node (r) [vector] {$\bm r$};& & \node (m1) [function] {$\bm\odot$}; & &
        \node (a1) [function] {$\bm +$}; & &
        \node (log) [function] {log}; & &
        \node (wr1) [function] {matmul}; & & \\

      & \node (abs) [function] {abs}; & & & & & & & & & &
        \node (wi1) [function] {matmul};& & \node (plus) [function] {$\bm -$}; & &
        \node (exp) [function] {exp};\\

      \node (in) [vector] {$\bm x$}; & \node (p2) [point] {};& & \node (g) [weight] {$\bm g$}; &
      \node (clip) [function] {clip 0 1}; & & \node (minusone) [function] {1-g}; & & & &
      \node (wr) [weight] {$\bm W_r$}; & \node (wi) [weight] {$\bm W_i$}; & & & & &
      \node (mul) [function] {$\bm\odot$}; & &
      \node (out) [vector] {$\bm y$};\\

      & \node (0pi) [function] {0:$\pi$}; & & & & & & & & & &
        \node (wi2) [function] {matmul};& & \node (minus) [function] {$\bm +$};& &
        \node (cos) [function] {cos};\\

      & & \node (k) [vector] {$\bm k$}; & & \node (m2) [function] {$\bm\odot$}; & & & & &
          \node (beforewr2) [point] {}; & 
        \node (wr2) [function] {matmul}; & & & \\
    };

    { [start chain]
      \chainin (in);
      \chainin (p2) [join];
      { [start branch=rbr]
        \chainin (abs) [join];
        \chainin (r) [join=by {vh path}];
        \chainin (m1) [join=by tip];
        \chainin (a1) [join=by tip];
        \chainin (log) [join=by tip];
        { [start branch=rbrup]
          \chainin (wr1) [join=by tip];
          \chainin (plus) [join=by {tip, hv path}];
        }
        { [start branch=rbrdown]
          \chainin (wi2) [join=by {vh path, tip}];
          \chainin (minus) [join=by tip];
        }
        \chainin (plus);
        \chainin (exp) [join=by tip];
        \chainin (mul) [join=by {hv path,tip}];
      }
      { [start branch=kbr]
        \chainin (0pi) [join];
        \chainin (k) [join=by {vh path}];
        \chainin (m2) [join=by tip];
        \chainin (beforewr2) [join];
        { [start branch=kbrdown]
          \chainin (wr2) [join=by tip];
          \chainin (minus) [join=by {hv path,tip}];
        }
        { [start branch=kbrup]
          \chainin (wi1) [join=by {vh path,tip}];
          \chainin (plus) [join=by tip];
        }
        \chainin (minus);
        \chainin (cos) [join=by tip];
        \chainin (mul) [join=by {hv path,tip}];
        \chainin (out) [join=by tip];
      }
    }

    { [start chain]
      \chainin (g);
      \chainin (clip) [join=by fat];
      \chainin (m1) [join=by tip];
    }

    { [start chain]
      \chainin (clip);
      \chainin (m2) [join=by tip];
    }

    { [start chain]
      \chainin (clip);
      \chainin (minusone) [join=by tip];
      \chainin (a1) [join=by tip];
    }

    { [start chain]
      \chainin (wr2);
      \chainin (wr) [join=by fat];
      \chainin (wr1) [join=by fat];
    }

    { [start chain]
      \chainin (wi2);
      \chainin (wi) [join=by fat];
      \chainin (wi1) [join=by fat];
    }

\end{tikzpicture}

}
    \captionof{figure}{NPU diagram. The NPU has a relevance gate $\bm g$
    (hatched background) in front of the input to the unit to prevent
    zero gradients.}
    \label{fig:gatednpu_diagram}
  \end{minipage}
\end{figure}

\subsection{The Relevance Gate -- NPU}%
\label{sub:npu}

\begin{figure}
  \centering
  \includegraphics[width=.6\textwidth]{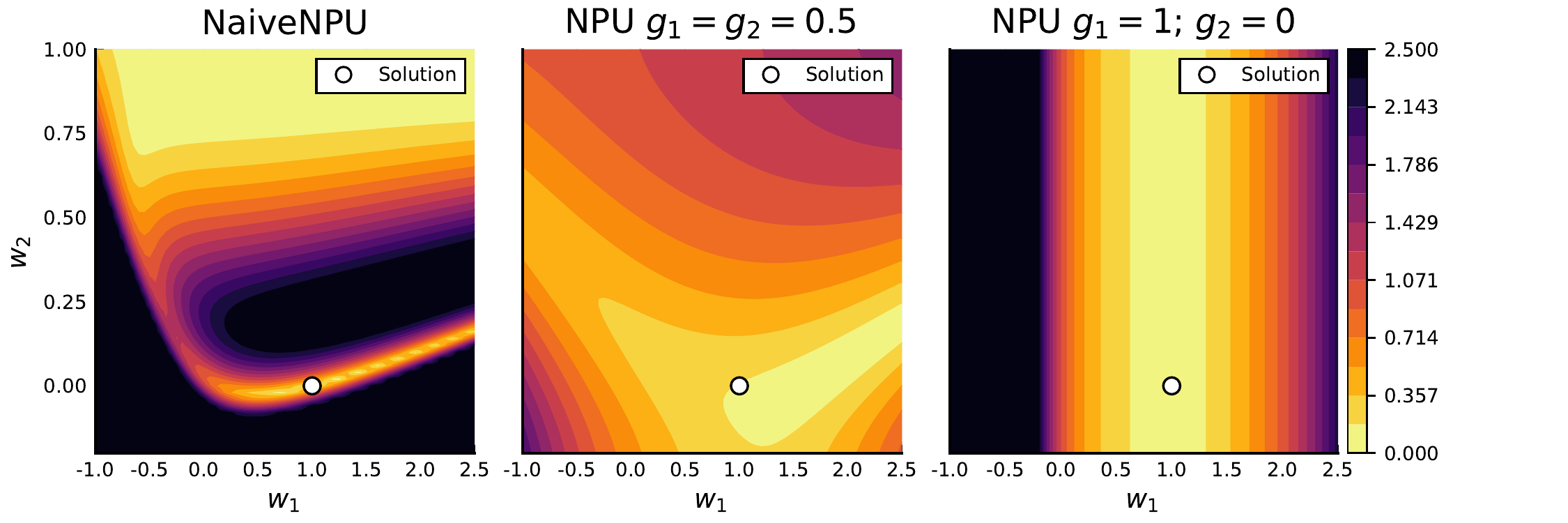}
  \caption{Gradient norm $\mathcal G$ of NaiveNPU and NPU for the task of learning
  the identity on $x_1$ (black areas are beyond the color scale). Inputs and
  loss are defined in Eq.~\ref{eq:grad_surf}. The correct solution is
  $w_1=1$ and $w_2=0$.  The NaiveNPU has a large zero gradient region for
  $w_2>0.75$, while the NPU's surface is much more informative. The gates
  in the central plot are fixed at $g_1=g_2=0.5$ which corresponds to the initial
  gate parameters. During training they adjust as needed, in this case to
  $g_1=1$ and $g_2=0$. $\bm W_i$ is set to zero in all plots.
  }%
  \label{fig:id_loss}
\end{figure}

The NaiveNPU has difficulties to converge on large scale tasks, and to reach
sparse results in cases where the input to a given row is small. We
demonstrate this on a toy example of learning the function
$f:\mathbb{R}^2 \rightarrow \mathbb{R}$,
which is the identity on one of two inputs. The task is defined by the loss $\mathcal L$:
\begin{gather}
  \mathcal{L} = \sum_i |m(x_1,x_2) - f(x_1,x_2)| = \sum_i |m(x_1,x_2) - x_{1,i}|,\nonumber\\
  \label{eq:grad_surf}
  \text{where } m = \naivenpu \text{ with } (\Wre,\Wim) \in \mathbb{R}^{1\times 2} \\
  \text{and } x_1 \sim \mathcal U(0,2), x_2 \sim \mathcal U(0,0.05). \nonumber
\end{gather}
The left plot in Fig.~\ref{fig:id_loss} depicts the gradient norm $\mathcal{G}$
\begin{equation}
  \mathcal{G}(\Wre) = \norm{\frac{\partial\mathcal{L}}{\partial\Wre}}_2
\end{equation}
of the NaiveNPU for a batch of two-dimensional inputs.
Even in this simple example, the gradient of the NaiveNPU is close to
zero in large parts of the parameter space. This can be explained as follows.
One row of NaiveNPU weights effectively raises each input to a power and
multiplies them: $x_1^{w_1} x_2^{w_2} \dots x_n^{w_n}$. If a single input $x_i$
is constantly close to zero (i.e. irrelevant), the whole row will be zero, no
matter what its weights are and the gradient information on all other weights
is lost.  Therefore, we introduce a gate on the input of our layer that can turn
irrelevant inputs into 1s.
A diagram of the NPU is shown in Fig.~\ref{fig:gatednpu_diagram}.

\begin{definition}[{\bf NPU}]
  The NPU extends the NaiveNPU by the relevance gate $\bm g$ on the input $\bm x$.
   \begin{gather}
    \label{eq:gatednpu_def}
    \bm y = \exp(\Wre \log\bm r - \pi\Wim\bm k) 
          \odot \cos(\Wim\log \bm r + \pi\Wre\bm k), \text{ where } \\
    \bm r = \bm{\hat g} \odot (|\bm x|+\epsilon) + (1-\bm{\hat g}),
    \quad
    k_i = \begin{cases}
       0  & x_i \geq 0 \\
      \hat g_i & x_i < 0
    \end{cases},
    \quad
    \hat g_i = \min(\max(g_i,0),1),
  \end{gather}
  with inputs $\bm x$, and learnt parameters $\Wre$, $\Wim$ and $\bm g$.
\end{definition}

The central plot of Fig.~\ref{fig:id_loss} shows the gradient norm $\mathcal G$ of the NPU
on the identity task with its initial gate setting of $g_1=g_2=0.5$. The
large zero-gradient region of the NaiveNPU is gone. The last plot shows the
same for $g_1=1$ and $g_2=0$, which corresponds to the correct gates at
the end of NPU training. The gradient is independent of $w_2$, which
means that it can easily be pruned by a simple regularization such as $L_1$. In
Sec.~\ref{sub:large_scale_arithmetic_task} we show how important the relevance
gating mechanism is for the convergence and sparsity of large models.  Sparsity
is especially important in order to use the NPU as a transparent model.

\textbf{Initialization}%
\label{sub:initialization}
We recommend initializing the NPU with a Glorot Uniform distribution on the
real weights $\Wre$. The imaginary weights $\Wim$ can be initialized to zeros,
so they will only be used where necessary, and the gate $\bm g$ with 0.5, so
the NPU can choose to output 1.

\begin{definition}[\textbf{RealNPU}]
  In many practical tasks, such as multiplication or division, the final value
  of $\bm W_i$ should be equal to zero. We will denote NPU with removed
  parameters for the imaginary part as RealNPU and study the impact of this
  change on convergence in Sec.~\ref{sec:experiments}.
\end{definition}

\section{Experiments}%
\label{sec:experiments}

In Sec.~\ref{sub:fractional_sir_model}, we show how the NPU can help to build
better NODE models. Additionally, we use the RealNPU as a highly transparent
model, from which we can directly recover the generating equation of an ODE
containing fractional powers.  Subsequent
Secs.~\ref{sub:simple_arithmetic_task} and \ref{sub:large_scale_arithmetic_task}
compare the NPU to prior art (NALU and NMU) on arithmetic tasks typically used
to benchmark arithmetic units.

\subsection{A Step Towards Equation Discovery of an Epidemiological Model}%
\label{sub:fractional_sir_model}

Data-driven models such as \emph{SINDy}~\citep{champion_data-driven_2019} or
\emph{Neural Ordinary Differential Equations} (NODE, \cite{chen_neural_2019})
are used more and more in scientific applications.
Recently, \emph{Universal Differential Equations} (UDEs,
\cite{rackauckas_universal_2020}) were introduced which aim to combine
data-driven models with physically informed differential equations to maximize
interpretability/explainability of the resulting models.

If an ODE model is composed of dense layers, its direct interpretation is
problematic and has to be performed retrospectively. The class of models based on
SINDy is transparent by design, however it can only provide explanation
within a linear combination of predefined set of basis functions. Thus, it
cannot learn models with unknown fractional powers. 
With this experiment we aim to show that the NPU can \emph{potentially} discover
exact ODE models.\footnote{The demonstration given here is not intended to be
used in practice. For real-world predictions in such a sensitive area much more
post-processing is needed to ensure safe predictions.}
An example of an ODE that contains powers is a modification of the well-known
epidemiological SIR model \citep{kermack_contribution_1927} to fractional powers
(fSIR, \cite{taghvaei_fractional_2020}), which was shown to be a beneficial
modification for modelling the COVID-19 outbreak.  The SIR model is
built from three variables: $S$ (susceptible), $I$ (infectious), and $R$
(recovered/removed).  Arguably the most important part of the model is the
transmission rate $r$, which is typically taken to be proportional to the
product of $S$ and $I$.  \citet{taghvaei_fractional_2020} argue that,
especially in the initial phase of an epidemic, the boundary areas of infected
and susceptible cells scale with a fractional power, which leads to
Eq.~\ref{eq:fracsir_r}:
\begin{align}
  \label{eq:fracsir}
  \frac{dS}{dt} &= -r(t) + \eta R(t), &
  \frac{dI}{dt} &=  r(t) - \alpha I(t), &
  \frac{dR}{dt} &= \alpha I(t) - \eta R(t),\\
  \label{eq:fracsir_r}
  & & r(t) &= \beta I(t)^\gamma S(t)^\kappa,
\end{align}

\begin{figure}
  \centering
  \resizebox{.9\textwidth}{!}{\input{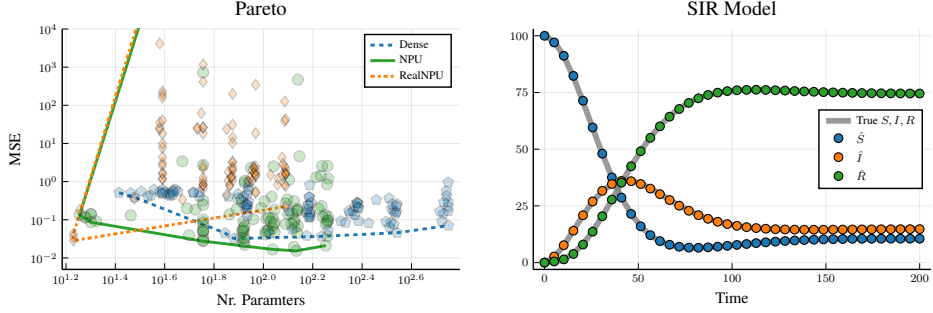}}
  \caption{Pareto fronts of the dense network, NPU, and RealNPU. The
  NPU reaches solutions with lower MSE and fewer parameters than the dense
  net. The RealNPU mostly yields worse results than the NPU. Just in a few
  cases it converges to very sparse models with good MSE.}%
  \label{fig:pareto_sir}
\end{figure}

\begin{figure}
  \centering
  \includegraphics[width=.8\linewidth]{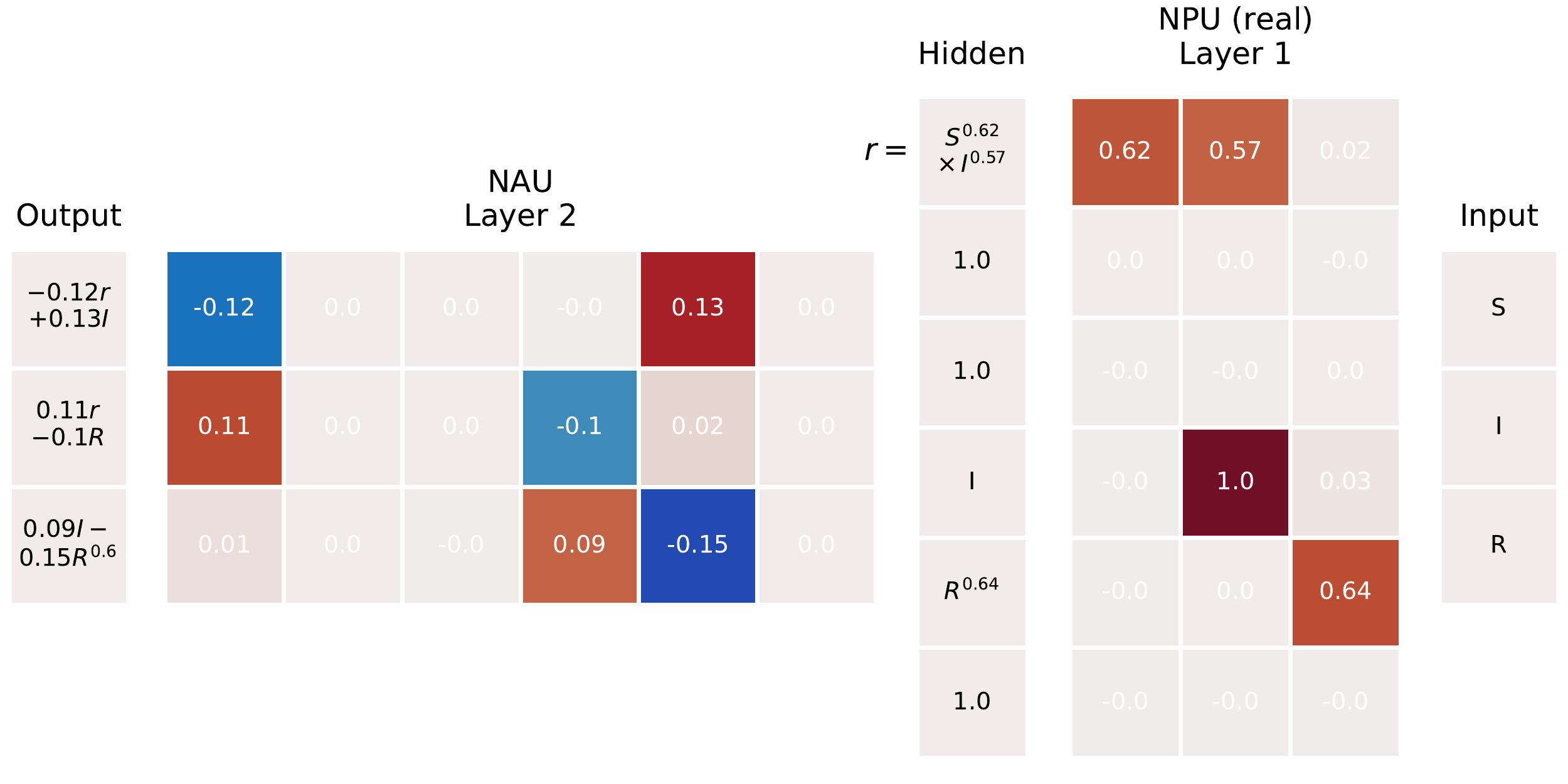}
  \captionof{figure}{Visualization of the best RealNPU. Reading from right
  to left, it takes the SIR variables as an input, then applies the NPU and
  the NAU.  It correctly identifies $r$ as a fractional product in the NPU,
  and gets the rest of the fSIR parameters almost right in the NAU.}%
  \label{fig:sir_gatednpu_modelps}
\end{figure}

We have numerically simulated one realization of the fSIR model with the
parameters $\alpha=0.05$, $\beta=0.06$, $\eta=0.01$, $\gamma=\kappa=0.5$, in 40
time steps that are equally spaced in the time interval $T=(0,200)$, such that
the training data $\bm X = [S_t,I_t,R_t]_{t=1}^{40}$
contains one time series each for $S$, $I$, and $R$.  The initial conditions
$\bm u_0=[S_0,I_0,R_0]$ are set to $S_0=100$, $I_0=0.01$, and $R_0=0$, as shown in Figure
\ref{fig:pareto_sir} (right).
We fit the data with three
different NODEs composed of different model types: a dense network, the NPU, and
the RealNPU. An exemplary model is: $\npu=\chain(\npu(3,h),\nau(h,3))$ with
variable hidden size $h$.  The detailed models are defined in
Tab.~\ref{tab:fsir_models}. The training objective is the loss $\mathcal{L}$
with $L_1$ regularization.
\begin{align}
  \label{eq:sir_loss}
  \mathcal{L} = \text{MSE}(\bm X,\text{NODE}_\theta(\bm u_0)) + \beta||\bm\theta||_1.
\end{align}
We train each model for 3000 steps with the ADAM optimizer and a learning rate
of 0.005, and subsequently with LBFGS until convergence (or for maximum 1000
steps).  For each model type, we run a small grid search to build a Pareto front
with $h\in \{6,9,12,15,20\}$ and $\beta\in\{0,0.01,0.1,1\}$, where each
hyper-parameter pair is run five times. The resulting Pareto front is shown on
the left of Fig.~\ref{fig:pareto_sir}. The NPU reaches much sparser and
better solutions than the dense network.
The RealNPU has problems to converge in the majority of cases, however, 
there are a few models in the bottom left that
reach a very low MSE and have very few parameters. The best of these models is
shown in Fig.~\ref{fig:sir_gatednpu_modelps}.  It looks strikingly similar to
the fSIR model in matrix form:
\renewcommand*{\arraystretch}{1.3}
\begin{equation}
  \begin{bmatrix}
    \dot S \\ \dot I \\ \dot R
  \end{bmatrix}
  =
  \begin{bmatrix}
    -\beta & 0 & \eta \\
    \beta & -\alpha & 0 \\
    0 & \alpha & \eta
  \end{bmatrix}
  \begin{bmatrix}
    I^\gamma S^\kappa \\ I \\ R
  \end{bmatrix}.
\end{equation}
Reading Fig.~\ref{fig:sir_gatednpu_modelps} from right to left, we can extract
the ODE that the RealNPU represents.  The first hidden variable correctly
identified the transmission rate as a product of two fractional powers $r =
I^\gamma S^\kappa$ with $\kappa=0.57$ and $\gamma=0.62$, which is close to the
true values $\gamma=\kappa=0.5$. The second, third, and the last hidden variable
were found to be irrelevant (the relevance gate returns 1). The fourth hidden
variable is a selector of the second input $I$, and the fifth hidden variable
is selector of a power of $R$, $R^{0.64}$. In the second layer, the NAU combines
the correct hidden outputs from the NPU such that $\dot S$ is composed of the
negative transmission rate $r$ and positive $R$.  $\dot I$ and $\dot R$ are
also composed of the correct hidden variables, with the parameters
$\alpha,\beta,\eta$ being not far off from the truth.  We conclude that even
with this very naive approach, the RealNPU can recover something close to the
true fractional SIR model.
\\
In summary, the NPU can work well in sequential tasks, and we have shown
that we can reach highly transparent results with the RealNPU, but in practice,
using the RealNPU might be difficult due to its lower success rate. With a more
elaborate analysis, it should be possible to reach the same solutions with the
full NPU and e.g. a strong regularization of its imaginary parameters.

\subsection{Simple Arithmetic Task}%
\label{sub:simple_arithmetic_task}

\begin{figure}
  \centering
  \includegraphics[width=\linewidth]{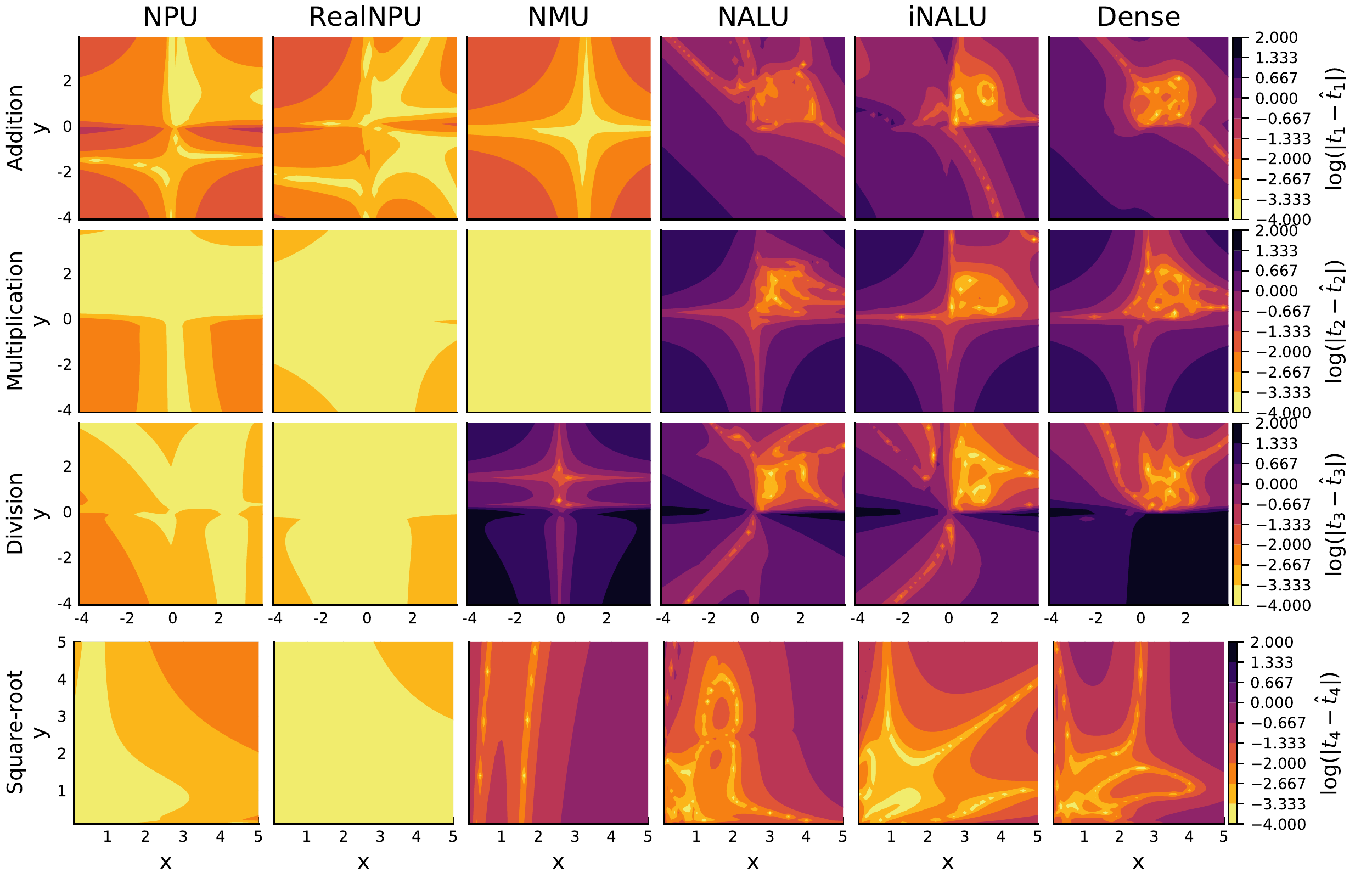}
  \caption{Comparison of extrapolation quality of different models learning Eq.~\ref{eq:simple_f}. Each
  column represents the best model of 20 runs that were trained on the range
  $\mathcal{U}(0.1,2)$.  Lighter color implies lower error.}%
  \label{fig:simple_err}
\end{figure}

In this experiment we compare six different layers (NPU, RealNPU, NMU, NALU,
iNALU, Dense) on a small problem with two inputs and four outputs. The
objective is to learn the function $f: \mathbb{R}^2
\rightarrow \mathbb{R}^4$ with a standard MSE loss:
\begin{align}
  \label{eq:simple_f}
  f(x,y) &= (x+y,\, xy,\, x/y,\, \sqrt{x} \text{  })^T \eqqcolon \bm t, \\
  \mathcal L &= \frac{1}{4} \sum_{i=1}^{4}\left(\text{model}(x,y)_i - f(x,y)_i\right)
             = \text{MSE}(\bm{\hat t}, \bm t).
\end{align}
Learning the function $f$ includes not only learning the correct arithmetic operation,
but also to separate them cleanly, which tests the gating mechanisms of the layers.
Each model has two layers with a hidden dimension $h$.  E.g. the NPU model is
defined by $\npu=\chain(\npu(2,h=6),\nau(h=6,4))$.  The remaining models that are
used in the tables and plots are given in Tab.~\ref{tab:models_simple_task}.
To obtain valid results in case of division we train on positive, non-zero
inputs, but test on negative, non-zero numbers (except for test inputs to the
square-root):
\begin{align}
  \label{eq:simple_train}
  (x_{\text{train}}, y_{\text{train}}) &\sim \mathcal{U}(0.1,2) &
  (x_{\text{test}}, y_{\text{test}}) &\sim \mathcal{R}(\text{-4.1:0.2:4}) &
  (x_{\text{test,\,sqrt}}, y_{\text{test,\,sqrt}}) &\sim \mathcal{R}(\text{0.1:0.1:4})
\end{align}
where $\mathcal{R}$ denotes a \emph{range} with start, step, and end.
We train each model for 20\,000 steps with the ADAM optimizer, a learning rate of
0.001, and a batch size of 100. The input samples are generated on the fly
during training. Fig.~\ref{fig:simple_err} shows the error surface of the best
of 20 models on each task. Tab.~\ref{tab:simple_err} lists the corresponding
averaged testing errors of all 20 models.

Both NPUs successfully learn ($+,\times,\div,\sqrt{\cdot}$) and clearly
outperform NALU and iNALU on all tasks.
Surprisingly, the NALU has problems extrapolating in this task, which as
\cite{schlor_inalu_2020} suggest, might be due to its gating mechanism.
The NPUs are on par with the NMU for
($+$), but the NMU is better at ($\times$) due to its inductive bias. The NMU
cannot learn ($\div,\sqrt{\cdot}$). The fact that the RealNPU performs slightly better
than the NPU indicates that the task is easy enough to not require the
imaginary parameters to help convergence. In such a case, the RealNPU
generalizes better because it corresponds to the task it
is trying to learn.

\subsection{Large Scale Arithmetic Task}%
\label{sub:large_scale_arithmetic_task}

One of the most important properties of a layer in a neural network is its
ability to scale. With the large scale arithmetic task we show that the NPU
works reliably on many-input tasks that are heavily over-parametrized.  In this
section we compare NALU, NMU, NPU, RealNPU, and the NaiveNPU on a task that is identical
to the `arithmetic task' that \cite{madsen_neural_2020} and
\cite{trask_neural_2018} analyse as well.
The goal is to sum two subsets of a 100 dimensional vector and
apply an operation (like $\times$) to the two summed subsets.  The dataset
generation is defined in the set of Eq.~\ref{eq:large_scale_def}, with the
parameters from Tab.~\ref{tab:dataset_params_large_scale}.
\begin{align}
  \label{eq:large_scale_def}
  a = \sum_{i=s_{1,\text{start}}}^{s_{1,\text{end}}} x_i, &&
  b = \sum_{i=s_{2,\text{start}}}^{s_{2,\text{end}}} x_i, &&
  y_{\text{add}} = a + b, && y_{\text{mul}} = a \times b, &&
  y_{\text{div}} = 1 / a, && y_{\text{sqrt}} = \sqrt{a},
\end{align}
where starting and ending values $s_{i,\text{start}},s_{i,\text{end}}$ of the
summations are chosen such that $a$ and $b$ come from subsets of the input
vector $\bm x$ with a given overlap.  The training objective is standard MSE,
regularized with $L_1$:
\begin{equation}
  \mathcal{L} = \text{MSE}(\text{model}(\bm x), y) + \beta\norm{\bm\theta}_1,
\end{equation}
where $\beta$ is scheduled to be low in the beginning of training and stronger
towards the end.  Specifics of the used models and their hyper-parameters are defined in
Tab.~\ref{tab:models_large_task} \& \ref{tab:training_large_scale}.  \cite{madsen_neural_2020} perform an
extensive analysis of this task with different subset and overlap ratios,
varying model and input sizes, and much more, establishing that the combination
of NAU/NMU outperforms the NALU. We focus on the
comparison of NPU, RealNPU, NMU, and NALU on the default parameters of
\cite{madsen_neural_2020} which sets the subset ratio to 0.5 and the overlap
ratio to 0.25 (details in Tab.~\ref{tab:dataset_params_large_scale}).
We include the NaiveNPU (without the relevance gate) to show
how important the gating mechanism is for both sparsity and overall
performance.

\begin{figure}
  \centering
  \resizebox{\textwidth}{!}{\input{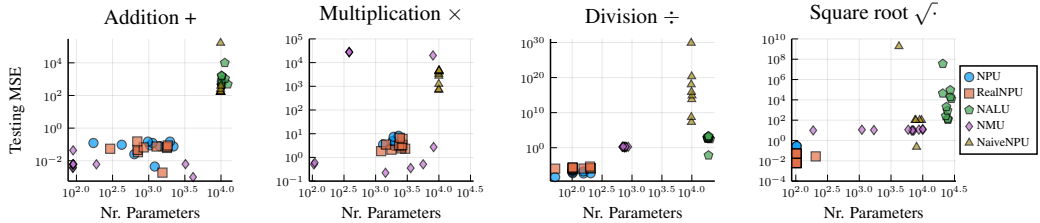}}
  \caption{Testing MSE over number of non-zero parameters ($w_i > 0.001$) of the large scale
  arithmetic task. The NMU outperforms the NPU on its
  native tasks, addition and multiplication. The NPU is the best at division
  and square-root.  The NaiveNPU without the relevance gate is far off, because
  it does not have the necessary gradient signal to converge, as discussed in
  Sec.~\ref{sub:npu}}%
  \label{fig:pareto}
\end{figure}

Fig.~\ref{fig:pareto} plots testing errors over the number of non-zero
parameters for all models and tasks.  The addition plot shows that NMU, NPU,
and RealNPU successfully learn and extrapolate on ($+$) with the NMU converging
to the sparsest and most accurate models.
On ($\times$), the best NMU models outperform the NPU and RealNPU, but some
NMUs do not converge at all. The testing MSE of the NALU is so large that it is
excluded from the plot.
On ($\div,\sqrt{\cdot}$) the NPU clearly outperforms all other layers
in MSE and sparsity. Generally, the difference between the NaiveNPU and the other NPUs
is huge and demonstrates how important the relevance gate is both
for convergence and sparsity. The NPUs with relevance gates effectively convert
irrelevant inputs to 1s, while the NaiveNPU is stuck on the zero gradient
plateau.

\begin{table}
  \centering
  \caption{Testing errors of the large scale arithmetic task. Each value is
  obtained by computing median (and median absolute deviation) of 10 runs.}
  \label{tab:arithmetic100_val}
  \small
  \begin{tabular}{lccccc}
\toprule
Task & NPU & RealNPU & NALU & NMU & NaiveNPU\\
\midrule
$+$  & 0.092 $\pm$ 0.031 & 0.063 $\pm$ 0.014 &740.0 $\pm$ 330.0 & \textbf{0.00602 $\pm$ 0.00019} & 161.65 $\pm$ 0.11 \\
$\times$ & 4.28 $\pm$ 0.9 & 3.09 $\pm$ 0.74 &2.9e83 $\pm$ 2.9e83 & \textbf{1.7 $\pm$ 1.4} & 3750.0 $\pm$ 870.0 \\
$\div$  & \textbf{1.0e-7 $\pm$ 1.0e-7} & 1.4e-6 $\pm$ 4.0e-7 &530.0 $\pm$ 200.0 & 1.622 $\pm$ 0.081 & 5.4e17 $\pm$ 5.4e17 \\
$\sqrt{\cdot}$ & 0.054 $\pm$ 0.0078 & \textbf{0.017 $\pm$ 0.011} &7300.0 $\pm$ 7200.0 & 10.96 $\pm$ 0.89 & 9.3e8 $\pm$ 9.3e8 \\
\bottomrule
\end{tabular}

\end{table}

\section{Conclusion}%
\label{sec:conclusion}

We introduced the \emph{Neural Power Unit} which addresses the deficiencies
of current arithmetic units: it can learn multiplication, division, and arbitrary power functions for
positive, negative, and small numbers. We showed that the NPU outperforms its
main competitor (NALU) and reaches performance that is on par with the
multiplication specialist NMU (Sec.~\ref{sub:simple_arithmetic_task} and
\ref{sub:large_scale_arithmetic_task}).\\
Additionally, we have demonstrated that the NPU converges consistently, even on
sequential tasks. The RealNPU can be used as a transparent model that
is capable of recovering the governing equations of dynamical systems purely
from the data (Sec.~\ref{sub:fractional_sir_model}).

\section*{Broader Impact}%
\label{sec:statement_of_broader_impact}

Current neural network architectures are often perceived as black box models
that are difficult to explain or interpret. This becomes highly problematic if
ML models are involved in high stakes decisions in e.g. criminal justice,
healthcare, or control systems.  With the NPU, we hope to contribute to the
broad topic of interpretable machine learning, with a focus on scientific
applications.
Additionally, learning to abstract (mathematical) ideas and extrapolating is a
fundamental goal that might contribute to more reliable machine learning
systems.

However, the inductive biases that are used to increase transparency in the NPU
can cause the model to ignore subgroups in the data. This is not an issue for learning
arithmetic operations, but could lead to biased models in more general use cases.

The methodology presented in the experiments in Sec.~\ref{sub:fractional_sir_model}
is not indented to be used for real-world epidemiological predictions.
They are merely demonstrating that the NPU can learn an ODE with fractional powers.
For an application of the NPU much more post-processing has to be done to ensure
the reliability of the results.

\section*{Acknowledgements and Disclosure of Funding}%
\label{sec:acknowledgements}

The research presented in this work has been supported by the Grant Agency of
Czech Republic no. 18-21409S. The authors also acknowledge the support of the
OP VVV MEYS funded project CZ.02.1.01/0.0/0.0/16\_019/0000765 ``Research Center
for Informatics''.

We thank the authors of the Julia packages
Flux.jl~\citep{innes_fashionable_2018} and
DifferentialEquations.jl~\citep{rackauckas_differentialequationsjl_2017}).

\bibliographystyle{plainnat}
\bibliography{refs}

\newpage
\appendix
\section*{Appendix}%
\label{sec:appendix}

\setcounter{table}{0}
\renewcommand{\thetable}{A\arabic{table}}

\begin{table}[h]
  \centering
  \caption{Model definitions for the fSIR task.}
  \label{tab:fsir_models}
  \begin{tabular}{llll}
    \toprule
    Model & Layer 1 & Layer 2 & Layer 3 \\
    \midrule
    NPU & $\npu(3,\,h)$ & $\nau(h,\,3)$ & -- \\
    NPU & NPU$_{\text{real}}(3,\,h)$ & $\nau(h,\,3)$ & -- \\
    Dense & $\dense(2,\,h,\sigma)$ & $\dense(h,\,h,\,\sigma)$ & $\dense(h,\,3)$ \\
    \bottomrule
  \end{tabular}
\end{table}

\begin{table}[h]
  \centering
  \caption{Testing error on the simple arithmetic task for the different models
  (i.e. mean of each heatmap in Fig.~\ref{fig:simple_err}). Each value is
  obtained by computing median (and median absolute deviation) of the error of
  20 models.}
  \label{tab:simple_err}
  \small
  \begin{tabular}{lcccccc}
\toprule
Task & NPU & RealNPU & NMU & NALU & iNALU & Dense\\
\midrule
$+$  & 0.2 $\pm$ 0.11 & \textbf{0.08 $\pm$ 0.021} & 0.2 $\pm$ 0.18 & 2.69 $\pm$ 0.22 & 2.18 $\pm$ 0.13 & 2.103 $\pm$ 0.04 \\
$\times$ & 0.37 $\pm$ 0.23 & 0.066 $\pm$ 0.026 & \textbf{0.005 $\pm$ 0.004} & 4.55 $\pm$ 0.2 & 3.453 $\pm$ 0.065 & 3.546 $\pm$ 0.035 \\
$\div$  & 0.23 $\pm$ 0.13 & \textbf{0.085 $\pm$ 0.038} & 11.399 $\pm$ 0.035 & 3.33 $\pm$ 0.18 & 2.54 $\pm$ 0.26 & 14.16 $\pm$ 0.23 \\
$\sqrt{\cdot}$ & 0.031 $\pm$ 0.025 & \textbf{0.004 $\pm$ 0.001} & 0.16 $\pm$ 0.002 & 0.034 $\pm$ 0.006 & 0.049 $\pm$ 0.011 & 0.084 $\pm$ 0.007 \\
\bottomrule
\end{tabular}

\end{table}

\begin{table}[h]
  \centering
  \caption{Model definitions for the simple arithmetic task.}
  \label{tab:models_simple_task}
  \begin{tabular}{llll}
    \toprule
    Model & Layer 1 & Layer 2 & Layer 3 \\
    \midrule
    NPU & NAU(2,\,6) & NPU(6,\,2) & -- \\
    RealNPU & NAU(2,\,6) & RealNPU(6,\,2) & -- \\
    NMU & NAU(2,\,6) & NMU(6,\,2) & -- \\
    NALU & NALU(2,\,6) & NALU(6,\,2) & -- \\
    iNALU & iNALU(2,\,6) & iNALU(6,\,2) & -- \\
    Dense & Dense(2,\,10,\,$\sigma$) & Dense(10,\,10,\,$\sigma$) & Dense(10,\,2) \\
    \bottomrule
  \end{tabular}
\end{table}

\begin{table}[h]
  \centering
  \caption{Model definitions for the large scale arithmetic task.}
  \label{tab:models_large_task}
  \begin{tabular}{llll}
    \toprule
    Model & Layer 1 & Layer 2\\
    \midrule
    NPU & NAU(100,\,100) & NPU(100,\,1) \\
    NPU & NAU(100,\,100) & NPU(100,\,1) \\
    NMU & NAU(100,\,100) & NMU(100,\,1) \\
    NALU & NALU(100,\,100) & NALU(100,\,1) \\
    \bottomrule
  \end{tabular}
\end{table}

\begin{table}[h]
  \centering
  \caption{Dataset parameters for the large scale arithmetic task.}
  \label{tab:dataset_params_large_scale}
  \begin{tabular}{lccccc}
    \toprule
    Task & Input size & Subset ratio & Overlap ratio & Training range  & Validation range\\
    \midrule
    Add  & 100        & 0.5          & 0.25          & Sobol(-1,1)     & Sobol(-4,4) \\
    Mult & 100        & 0.5          & 0.25          & Sobol(-1,1)     & Sobol(-4,4)\\
    Div  & 100        & 0.5          & --            & Sobol(0,0.5)    & Sobol(-0.5,0.5) \\
    Sqrt & 100        & 0.5          & --            & Sobol(0,2)      & Sobol(0,4) \\
    \bottomrule
  \end{tabular}
\end{table}

\begin{table}[h]
  \centering
  \caption{Training parameters for the large scale arithmetic task. The
  $\beta$-parameters define the stepwise exponential growth of the $L_1$
  regularization with start, step, growth, and end.}
  \label{tab:training_large_scale}
  \begin{tabular}{lcccccc}
    \toprule
    Task & Learning rate & Iterations & $\beta_{\text{start}}$ &
      $\beta_{\text{end}}$ & $\beta_{\text{step}}$  & $\beta_{\text{growth}}$ \\
    \midrule
    Add  & 1e-2 & 1e5 & 1e-5 & 1e-4 & 10\,000 & 10 \\
    Mult & 5e-3 & 1e5 & 1e-5 & 1e-7 & 10\,000 & 10 \\
    Div  & 5e-3 & 1e5 & 1e-9 & 1e-7 & 10\,000 & 10 \\
    Sqrt & 5e-3 & 1e5 & 1e-6 & 1e-4 & 10\,000 & 10 \\
    \bottomrule
  \end{tabular}
\end{table}

\end{document}